\documentclass[10pt,twocolumn,letterpaper]{article}

\usepackage{wacv}
\usepackage{times}
\usepackage{epsfig}
\usepackage{graphicx}
\usepackage{amsmath}
\usepackage{amssymb}
\usepackage{booktabs}
\usepackage{subfigure}
\usepackage{diagbox}
\usepackage{color}
\usepackage{colortbl}
\definecolor{blue}{rgb}{0,0,1}

\usepackage{pifont}
\newcommand{\cmark}{\ding{51}}%
\newcommand{\xmark}{\ding{55}}%
\usepackage{multirow}
\makeatletter
\@namedef{ver@everyshi.sty}{}
\makeatother
\usepackage{pgfplots}
\usepackage[accsupp]{axessibility}

\definecolor{turquoise}{cmyk}{0.65,0,0.1,0.3}
\definecolor{purple}{rgb}{0.65,0,0.65}
\definecolor{dark_green}{rgb}{0, 0.5, 0}
\definecolor{orange}{rgb}{0.8, 0.6, 0.2}
\definecolor{red}{rgb}{0.9, 0.1, 0.1}
\definecolor{darkred}{rgb}{0.6, 0.1, 0.05}
\definecolor{blueish}{rgb}{0.0, 0.3, .6}
\definecolor{light_gray}{gray}{0.95}
\definecolor{pink}{rgb}{1, 0, 1}
\definecolor{greyblue}{rgb}{0.25, 0.25, 1}

\def\wacvPaperID{1861} 

\wacvalgorithmstrack   

\wacvfinalcopy 

\ifwacvfinal
\usepackage[breaklinks=true,bookmarks=false]{hyperref}
\else
\usepackage[pagebackref=true,breaklinks=true,colorlinks,bookmarks=false]{hyperref}
\fi

\begin{document}

\title{Language-free Training for Zero-shot Video Grounding}

\author{
Dahye Kim$^{1}$\quad\quad Jungin Park$^{1}$\quad\quad Jiyoung Lee$^{2}$\quad\quad Seongheon Park$^{1}$\quad\quad Kwanghoon Sohn$^{1,3}$\thanks{Corresponding author}\\
$^1$Yonsei University \quad 
$^2$NAVER AI Lab \quad 
$^3$Korea Institute of Science and Technology (KIST)\\ 
{\tt\small {\{dadaday, newrun, sam121796, khsohn\}}@yonsei.ac.kr} \quad
{\tt\small lee.j@navercorp.com}}


\maketitle
\thispagestyle{empty}

\begin{abstract}
    Given an untrimmed video and a language query depicting a specific temporal moment in the video, video grounding aims to localize the time interval by understanding the text and video simultaneously.
    One of the most challenging issues is an extremely time- and cost-consuming annotation collection, including video captions in a natural language form and their corresponding temporal regions.
    In this paper, we present a simple yet novel training framework for video grounding in the zero-shot setting, which learns a network with only video data without any annotation. 
    Inspired by the recent language-free paradigm, i.e. training without language data, we train the network without compelling the generation of fake (pseudo) text queries into a natural language form. 
    Specifically, we propose a method for learning a video grounding model by selecting a temporal interval as a hypothetical correct answer and considering the visual feature selected by our method in the interval as a language feature, with the help of the well-aligned visual-language space of CLIP.
   Extensive experiments demonstrate the prominence of our language-free training framework, outperforming the existing zero-shot video grounding method and even several weakly-supervised approaches with large margins on two standard datasets.

\end{abstract}


\section{Introduction}
    In our daily life, we surf, think, and learn through loads of videos.
    By extension, we wish to search for the information we want in the videos.
    Video grounding (also called video moment retrieval) with natural language query aims to help such video search by automatically localizing a temporal moment for various applications such as video surveillance~\cite{collins2000system} and smart video search~\cite{sivic2003video, snoek2009mediamill}.

    A major challenge of video grounding is the exorbitant cost of constructing time interval annotations aligned to a given text that is also collected. Although recent fully-supervised video grounding (FSVG) methods~\cite{liu2021context,soldan2021vlg} have shown remarkable performance on the limited size of datasets~\cite{gao2017tall, krishna2017dense}, there is still room for improvement with scale-up training.
    Especially in such a field, large-scale training data is required to cover numerous video domains (\eg, instructional videos, movies, and so forth). 
    However, building massive annotations as more billion scales like image-language datasets, such as LAION-5B~\cite{schuhmann2022laion}, in video scale is an impractical solution.

    \begin{figure}[t]
        \centering
        \subfigure[Video grounding]{\includegraphics[width=0.9\linewidth]{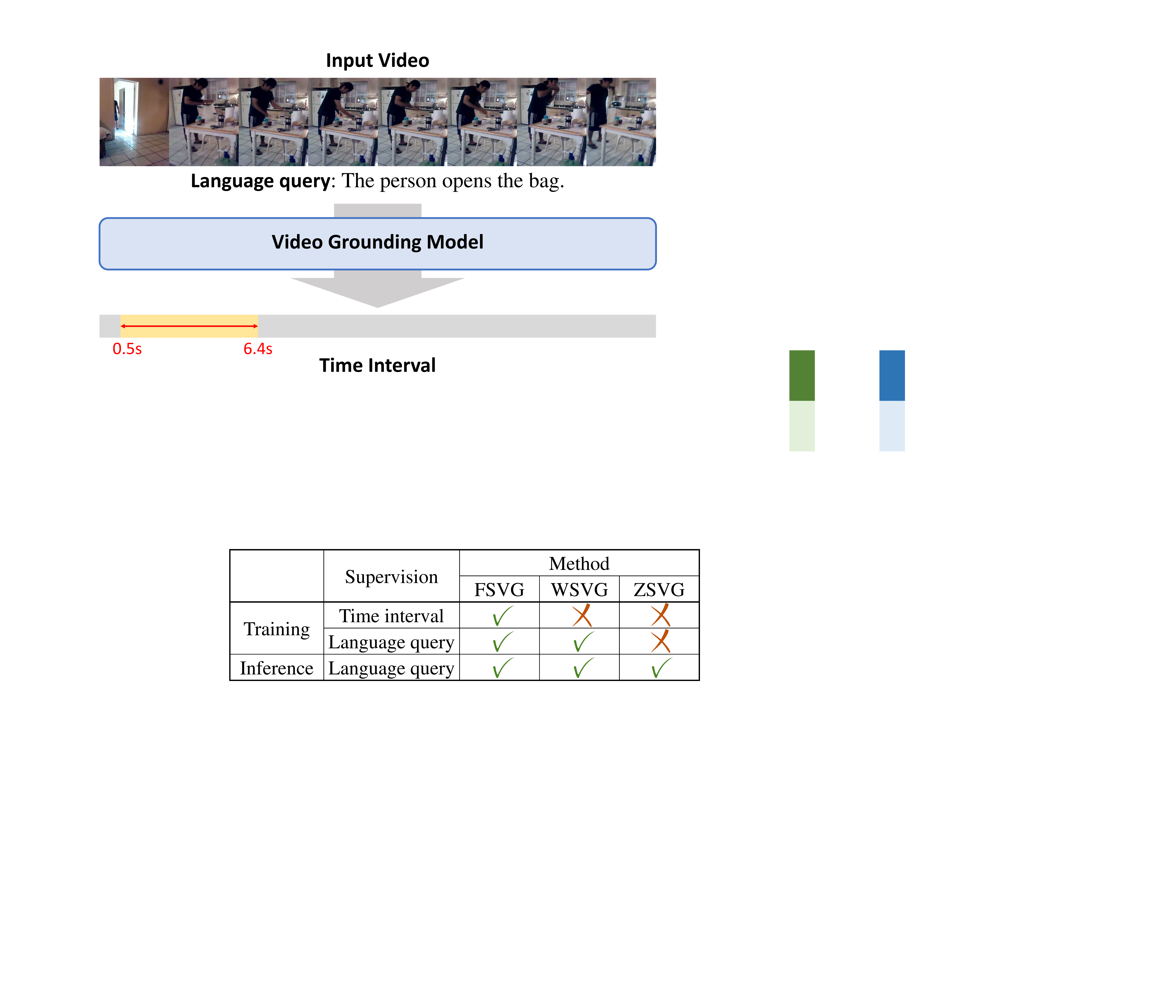}\label{fig:1}}\hfill \\
        \subfigure[Annotation types depending on the settings]{
        \resizebox{0.95\linewidth}{!}{
        \begin{tabular}{lccc}
        \toprule
        \multirow{2}{*}{\backslashbox{Setting}{~}} & \multicolumn{2}{c}{Training} & Test \\
        \cmidrule(lr){2-3} \cmidrule(lr){4-4}
        & Time interval & Language query & Language query \\
        \midrule
        FSVG & \textcolor{blueish}{\cmark} & \textcolor{blueish}{\cmark} & \textcolor{blueish}{\cmark}\\
        WSVG & \textcolor{darkred}{\xmark} & \textcolor{blueish}{\cmark} & \textcolor{blueish}{\cmark}\\
        ZSVG & \textcolor{darkred}{\xmark} & \textcolor{darkred}{\xmark} & \textcolor{blueish}{\cmark}\\
        \bottomrule
        \end{tabular}
        }
        }
        \caption{Given a video and a language query, video grounding aims to retrieve the time interval corresponding to the language query in the video. In this paper, we address the zero-shot video grounding (ZSVG) problem which is the most challenging setting and cannot use any annotations for training.
            }\label{fig:1}
    \end{figure}

    To address the burden of annotations, researchers have proposed weakly-supervised video grounding (WSVG) methods~\cite{gao2019wslln, lin2020weakly, mithun2019weakly} which use only coarse video-level descriptions for training. But they still require paired video-language data, showing limited applicability in the open world. Recently zero-shot video grounding (ZSVG) has been proposed in~\cite{nam2021zero}. As illustrated in~\figref{fig:1}, ZSVG utilizes only videos to learn the video grounding model in the training stage. To learn the localizing capability in a semi-supervised manner, \cite{nam2021zero} generates pseudo temporal event regions and corresponding pseudo sentence queries by examining noun-verb statistical co-occurrence patterns. However, pseudo sentences are built upon the composition of nouns and verbs (\eg, `flip person switch door'), which is naturally different from the form of natural language query (\eg, `person flipped the light switch near the door.'). Namely, contrived sentences with the simple composition of nouns and verbs break the structural and compositional generalization inherent in natural language that might harm the performance~\cite{li2022compositional}. 
    
    In this paper, we propose a novel language-free training framework for zero-shot video grounding. Our solution is to treat the visual feature as pseudo textual information while being flexible in responding to the act of forcing sentences to generate pseudo forced sentences. Specifically, we leverage an image-language pretraining model (\ie, CLIP~\cite{radford2021learning}) trained on large-scale web-collected data that have revealed a breakthrough in the multi-modal research field.
    We conjecture that text and visual features can replace each other without trouble in that CLIP provides a well-aligned visual-language semantic space.
    
    To this end, we first generate temporal proposals that contain meaningful events from a given untrimmed video.
    With the visual encoder of CLIP, visual features are extracted from all the frames in the proposal.
    Then our learnable selection transformer takes a dominant feature that has a role of the pseudo language feature in a video grounding model instead of generating a natural sentence from the proposal.
    Therefore, our method is free to generate high-quality natural language form from the proposal. Moreover, since the dominant visual feature is directly used for the pseudo textual feature, our method has no need to produce textual embedding from a pseudo text label, which is a time-consuming yet necessary step for the training of the previous method~\cite{nam2021zero}. 
    Finally, the whole model is learned to predict time intervals corresponding to pseudo sentence features with generated temporal proposals as ground-truth.
    Our contributions are summarized three-fold:
    \begin{itemize}
        \item We introduce a language-free framework for video grounding that can be an affordable solution to effectively reduce the annotation cost.
        \item We validate the applicability of the pretrained visual-language model to the video-language task by providing extensive experimental analysis.
        \item Our language-free training framework outperforms the existing method, achieving state-of-the-art performance, and even shows comparable performance with weakly-supervised approaches on the Charades-STA~\cite{gao2017tall} and ActivityNet Captions~\cite{krishna2017dense} datasets.
    
    \end{itemize}

\begin{figure*}[!t]
    \centering
        \subfigure[Training]
   {\includegraphics[width=0.53\linewidth]{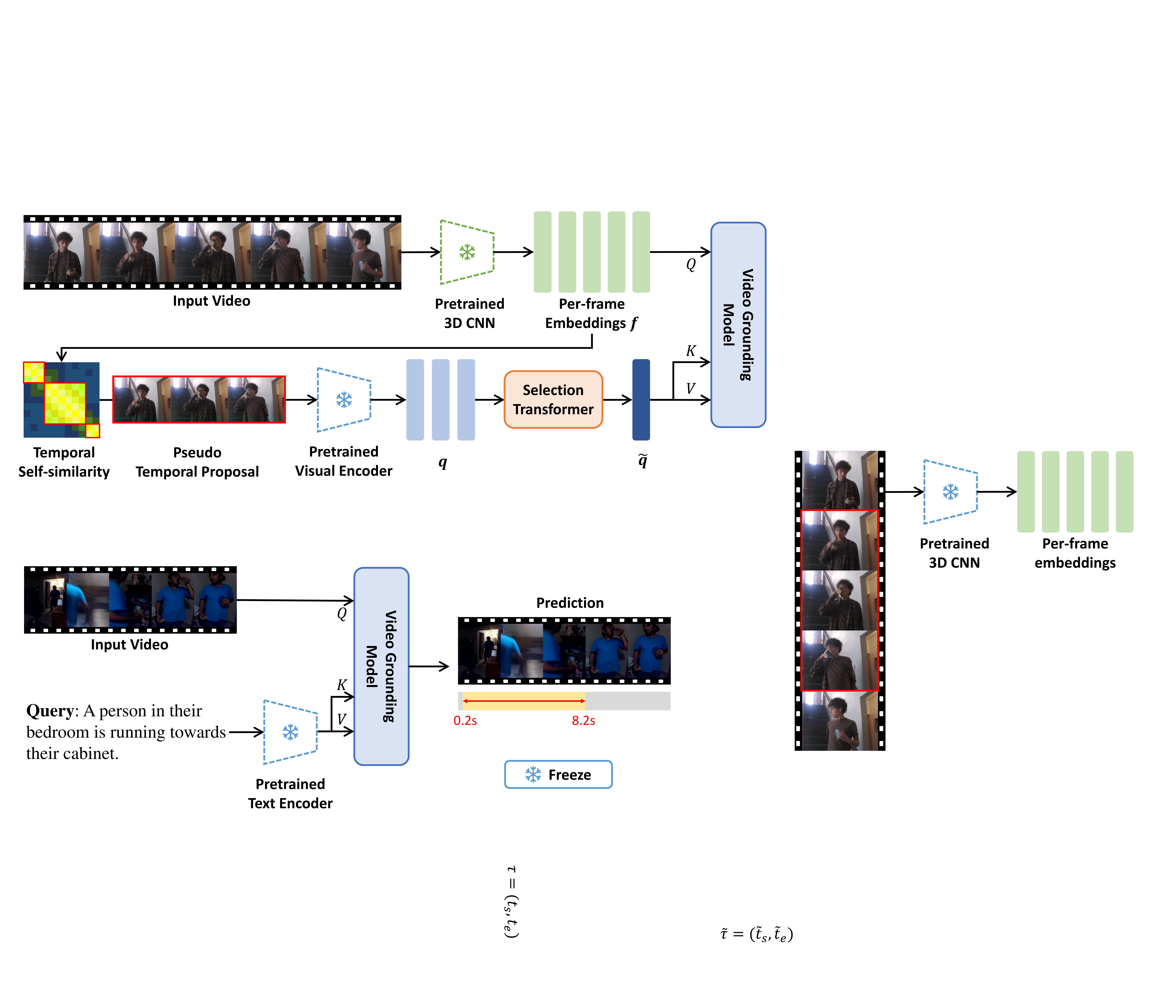}\label{fig:2-a}}\hfill
   \subfigure[Inference]{\includegraphics[width=0.46\linewidth]{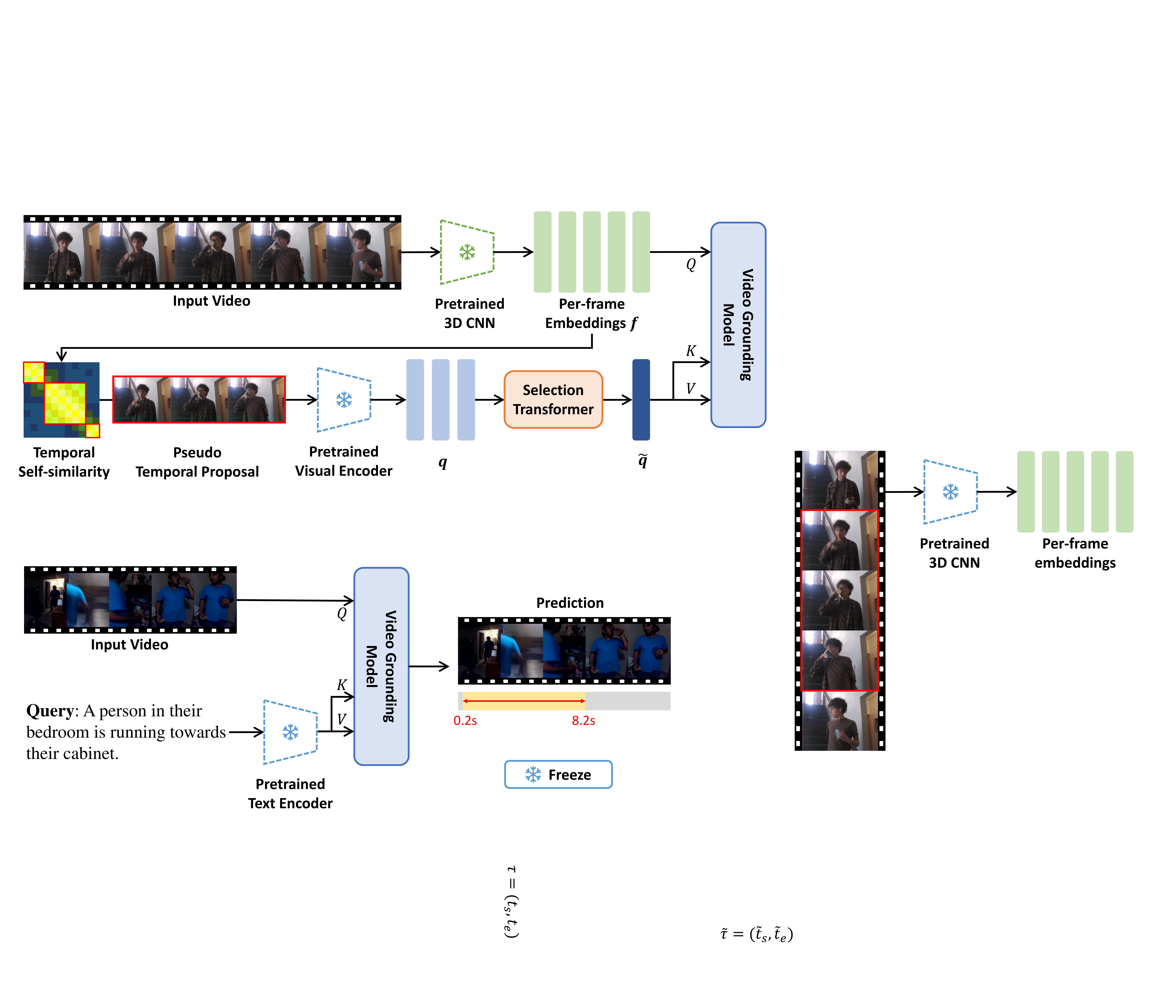}\label{fig:2-b}}\hfill
    	\caption{The overall framework of our language-free video grounding framework. In (a) training, we generate a pseudo temporal interval and corresponding pseudo language feature from the visual encoder of CLIP~\cite{radford2021learning} and selection transformer to train the video grounding model. In (b) inference (test) phase, we use the video grounding model only with text encoder of CLIP.
    }\label{fig:main}
\end{figure*} 

\section{Related Work}
\subsection{Video Grounding}
Video grounding is a recently proposed task~\cite{anne2017localizing, gao2017tall}, which aims to find the best moment in a video grounded on language queries.
Most of the existing methods followed fully-supervised setting~\cite{ding2021support, li2022compositional, liu2021context, mun2020local, rodriguez2020proposal,wang2021structured,yuan2019find, zeng2020dense, zhang2019man, zhang2020learning,zhao2021cascaded, zhou2021embracing} to model fine-grained semantic relations of video and language.
However, since such a setting requires precise annotations for the start and end timestamps, manual annotations of the temporal boundary were required, which also led to subjectivity across different annotators.

Weakly-supervised video grounding has been introduced to alleviate this burden. 
Existing works can be categorized into two groups. 
1) Multi-instance learning~\cite{dietterich1997solving} (MIL) based methods \cite{gao2019wslln, huang2021cross, ma2020vlanet, mithun2019weakly, tan2021logan, zhang2020counterfactual} utilized similarity scores by maximizing scores between positive samples and minimizing scores between negative samples. 
2) The reconstruction-based method~\cite{duan2018weakly,lin2020weakly,song2020weakly,  yang2021local,   zheng2022weakly} used the assumption that the video segment that best reconstructs the text query is close to the ground-truth.

However, while weakly-supervised approaches were successful in lowering the cost of temporal annotation, the cost of text query remains problematic.
Several works~\cite{liu2022unsupervised, nam2021zero} considered an unsupervised setting that does not access the paired annotations.
\cite{liu2022unsupervised} proposed a deep semantic clustering network, which first aggregates distinct semantic features from the whole query set and then generates pseudo labels to provide pseudo supervision for training. 
\cite{nam2021zero} generated pseudo labels of temporal boundaries and corresponding query sentences. 
They first utilized a temporal similarity matrix to find temporal event proposals, then used an off-the-shelf object detector and fine-tuned RoBERTa~\cite{liu2019roberta} to make a structure-less pseudo query. 
However, a structure-less pseudo query, especially composed of nouns and verbs, can be interpreted in several meanings, due to systematic compositionality~\cite{chomsky2009syntactic, fodor1988connectionism} of natural language. In addition, the existence of the uninformative word in the query makes it hard for the model to distinguish the exact meaning of what the query originally intended to mean.
Furthermore, inferred verbs from detected objects are loosely bonded in the sense that the verbs are not predicted directly from the video, which leads to the generation of inaccurate pseudo queries.

\subsection{Language-free Paradigm}
As recent trends shift from uni-modal learning to multi-modal learning, vision-language related tasks have attracted attention. 
Since the modality to be processed has doubled, it becomes difficult to obtain high-quality vision-language training pairs. 
Several works~\cite{nam2021zero, zhou2021lafite} proposed a so-called `language-free paradigm' to address this problem, which means training without language data in the vision-language tasks.

One line of work~\cite{feng2019unsupervised,jiang2022pseudo, laina2019towards,nam2021zero} presented a visual object-based approach, which utilizes an off-the-shelf object detector to make text-related pseudo labels based on detected objects. 
Unsupervised image captioning \cite{feng2019unsupervised, laina2019towards} utilized an object detector to explore visual concepts in an image from unpaired image and text information. 
Unsupervised visual grounding \cite{jiang2022pseudo} used detected objects as the first object proposals and then generated pseudo language queries with a pseudo-query generation module. 
Zero-shot video grounding \cite{nam2021zero} first detected objects from temporal event proposal as nouns, second utilized fine-tuned language model as verbs, and finally generated simplified sentence as pseudo query by composing nouns and verbs. 
However, the above methods heavily rely on the quality of recognized visual objects from object detectors, which has a large domain gap between the target dataset and the training dataset that the object detector has trained on. 
Furthermore, since the object categories were limited to the trained dataset, it was impossible to scale a wide variety of objects and rich expressions inherent in natural language~\cite{zhu2022prompt}.

Another line of work \cite{wang2022clip, zhou2021lafite,  zhu2022prompt} utilized well-aligned multi-modal semantic space of the pretrained visual-language model. 
\cite{zhu2022prompt} presented prompt-based learning method for unpaired image captioning model, which utilizes the vision-language alignment established by CLIP~\cite{radford2021learning}. 
\cite{wang2022clip,zhou2021lafite} proposed language-free text-to-image generation model using pretrained CLIP. 
Specifically, they generated a pseudo text feature directly from an image using CLIP, assuming that CLIP has learned image-text feature alignment in the joint embedding space. 
While we share the same spirit as the language-free text-to-image generation~\cite{zhou2021lafite}, our work is the first attempt to introduce language-free training for video grounding.

\section{Language-free Video Grounding}
\subsection{Problem Statement and Motivation}\label{sec:31}

    Given an untrimmed video and a language query, video grounding aims to localize a time interval (start and end time stamps) representing the content corresponding to the query. 
    In the zero-shot video grounding (ZSVL), the model is not allowed to access any language query and ground-truth time stamps during training. 
    To achieve this goal, the prior work~\cite{nam2021zero} generated a pseudo sentence query using a pretrained object detector and noun-verb statistics from text corpora.
    While they have successfully presented a baseline for zero-shot video grounding, there are still problems to be tackled:
    (1) they generated nouns for the pseudo query by heavily relying on the capacity of the pretrained object detector, which may have encoded inappropriate biases and has a limited number of object categories;
    (2) they trained the sentence query generation network and video grounding network separately, making the training procedure inefficient;
    (3) They assumed that simplified sentences consisting of nouns and verbs whose structural characteristics and compositional generalization inherent in natural language are ignored could be substitutes for natural language queries.
    
    To solve the aforementioned problems, we propose a language-free framework for ZSVL, which skips doubtful sentence generation for performance improvement and lightweight training.
    As shown in~\figref{fig:2-a}, the training pipeline of our framework is (1) constructing temporal proposals using a pretrained video encoder, (2) generating the pseudo language feature with a selection transformer among the frame-wise visual features from pretrained CLIP, and (3) training a video grounding model that will be used to inference.

\subsection{Temporal Proposal Generation}\label{sec:32}

As a first step toward language-free video grounding, we should generate temporal event proposals from a video that we regard as temporal ground-truth.
To detect events happening in the videos, we leverage a characteristic of visual similarity of consecutive frames.
Specifically, a temporal similarity matrix is constructed to segment videos where visually similar frames are activated. 
Since the temporal similarity matrix reflects the temporal structure of the given video~\cite{dwibedi2020counting, park2020sum, park2021b2a}, we utilize this information to find possible events occurring in the video.

Similar to~\cite{nam2021zero}, given raw video frames, we first extract the video feature from the sequence of segments using a pretrained video encoder $\mathcal{F}_v$. 
After obtaining extracted features $f$ which encode the temporal structure of each segment, we construct a self-similarity matrix $R$ of the given video as follows:
\begin{equation}
    R_{ij} = cos(f_i, f_j) = \frac{ f_i \cdot f_j}{\Vert f_i \Vert \Vert f_j \Vert},
\end{equation}
where $R_{ij}$ is the cosine similarity score between pairs of segment features $f_i$ and $f_j$.
Then we group the segments into $k$ dominant events by clustering the features using $k$-means algorithm.
Also, consecutive events are merged to deal with more complex events.

\subsection{Pseudo Language Feature Generation}\label{sec:33}
\paragraph{Candidates of a language feature.}
    To train the video grounding model, we need language queries corresponding to the generated temporal proposals.
    However, as mentioned in~\secref{sec:31}, creating a language query in a natural language form can neglect the natural property of the language and be erroneous and time-consuming. 
    Instead, motivated by the recent success of the zero-shot text-to-image generation~\cite{zhou2021lafite}, we employ the visual encoder of the vision-language model (\ie CLIP~\cite{radford2021learning}) trained on large-scale image-language data using contrastive loss.
    Since the visual and language features are well-aligned in the semantic space, we can use the visual feature as the pseudo language feature.

    Specifically, we randomly sample $N$ frames denoted by $\{v_j\}_{j=1}^{N}$ in each temporal proposal and encode frame-wise features using the pretrained visual encoder $\mathcal{F}_\text{img}$.
    Thus, a set of candidates $\mathbf{q}$ for the pseudo language feature
    \begin{equation}
         \mathbf{q} = \{q_1, ..., q_N\} = \{\mathcal{F}_\text{img}(v_1),...,\mathcal{F}_\text{img}(v_N)\},
    \end{equation}
    where ${q}_n$ denotes the visual feature corresponding to $v_n$.
    However, directly using the visual feature may not be enough to represent the real language features.
    To this end, we intentionally perturb the features from the pretrained visual encoder using random noise following \cite{zhou2021lafite}:
    \begin{gather}
        q_n \leftarrow q_n + \xi \epsilon ||q_n||_2 / ||\epsilon||_2, \\
        q_n \leftarrow q_n / ||q_n||_2,
    \end{gather}
    where $\epsilon \sim \mathcal{N}(0, \mathbf{I})$ is the Gaussian noise, $\xi > 0$ denotes a hyperparameter to control the degree of noise, and $||\cdot||_2$ is $L_2$ normalization. We note that ViT/B-32 of CLIP image encoder is used as $\mathcal{F}_\text{img}$ in this work.

\paragraph{Pseudo language feature selection.}
    Given encoded pseudo language feature candidates, we select a single dominant feature that is the most informative to represent the corresponding temporal proposal.
    While it is natural to encode temporal information in video-language tasks, we select the pseudo language feature without the temporal modeling.
    Our observation is that a single dominant visual feature can be more informative to represent the corresponding query for two main reasons:
    1) A video consists of consecutive frames that usually contain similar semantics from a continuous scene so that sampling a superior frame already contains important information of the video~\cite{buch2022revisiting,lei2021less};
    2) since a video is a collection of noisy frames due to the existence of background clutter or camera motion blur, the combination of sampled frames may contain uninformative information and be computationally inefficient.
    
    Moreover, inappropriate temporal modeling harms the vision-language semantic space, leading to an unreliable performance at inference time where a real language query is given.
    One alternative solution is leveraging a pretrained video-language model (\eg VideoCLIP~\cite{xu2021videoclip}).
    However, the video-language model is usually pretrained on a smaller number of video-language pairs (1.1M videos in VideoCLIP~\cite{xu2021videoclip}) than the visual-language model (400M image-text pair in original CLIP~\cite{radford2021learning}).
    Furthermore, video-language models typically require high computation and memory costs.
    Therefore, we insist that incorporating the visual-language model into our work efficiently leverages the confident visual-language semantic space. We will verify this observation in~\secref{sec:45}.
    
    Concretely, we formulate a selection transformer that has only simple two transformer layers for a frame selection process such that:
    \begin{equation}
        \texttt{ST}(\{ q_1, q_2, \ldots, q_N \}) \mapsto \tilde{q},
    \end{equation}
    where $\texttt{ST}$ is the selection transformer and $\tilde{q}$ denotes the pseudo language feature.
    To ensure the back-propagation of such transformer for an end-to-end training, we employ gumbel softmax similar to~\cite{buch2022revisiting}.

\begin{table*}[hbt!]
    \centering
    \begin{tabular}{l|c |  c c c c | c c c c c}
        \hline
        \multirow{2}{*}{Method}             & \multirow{2}{*}{Sup.} & \multicolumn{4}{c|}{Charades-STA}    & \multicolumn{5}{c}{ActivityNet Captions}      \\
                                            &     & R@0.3 & R@0.5 & R@0.7 & mIoU  & R@0.1 & R@0.3 & R@0.5 & R@0.7 & mIoU   \\  \hline \hline
        LGI~\cite{mun2020local}             & FS  & 72.96 & 59.46 & 35.48 & 51.38 & -     & 58.52 & 41.51 & 23.07 & 41.13    \\ 
        CTRL~\cite{gao2017tall}             & FS  &  -    & 21.42 & 7.15  & -     & 49.1  & 28.70 & 14.00 &  -    & 20.54    \\  \hline
        TGA~\cite{mithun2019weakly}         & WS  & 29.68 & 17.04 & 6.93  &  -    &   -   &   -   &   -   &   -   &   \\  
        CTF~\cite{chen2020look}             & WS  & 39.8  & 27.3  & 12.9  & 27.3  & 74.2  & 44.3  & 23.6  &   -   &  32.2     \\
        SCN~\cite{lin2020weakly}            & WS  & 42.96 & 23.58 & 9.97  &  -    & 74.48 & 47.23 & 29.22 &   -   &    -      \\ 
        WSTAN~\cite{wang2021weakly}         & WS  & 43.39 & 29.35 & 12.28 &  -    & 79.78 & 52.45 & 30.01 &   -   &    -   \\
        BAR~\cite{wu2020reinforcement}      & WS  & 44.97 & 27.04 & 12.23 &  -    &  -    & 49.03 & 30.73 &   -   &    -   \\
        MARN~\cite{song2020weakly}          & WS  & 48.55 & 31.94 & 14.81 &   -   &  -    & 47.01 & 29.95 &   -   &    -    \\
        CCL~\cite{zhang2020counterfactual}  & WS  &  -    & 33.21 & 15.68 &   -   &   -   & 50.12 & 31.07 &   -   &    -     \\
        LoGAN~\cite{tan2021logan}           & WS  & 51.67 & 34.68 & 14.54 &   -   &   -   &   -   &   -   &   -   &    -    \\
        CRM~\cite{huang2021cross}           & WS  & 53.66 & 34.76 & 16.37 &   -   & 81.61 & 55.26 & 32.19 &   -   &    -    \\
        VCA~\cite{wang2021visual}           & WS  & 58.58 & 38.13 & 19.57 & 38.49 & 67.96 & 50.45 & 31.00 &   -   &   33.15   \\
        LCNet~\cite{yang2021local}          & WS  & 59.60 & 39.19 & 18.87 & 38.94 & 78.58 & 48.49 & 26.33 &   -   &   34.29  \\
        RTBPN~\cite{zhang2020regularized}   & WS  & 60.04 & 32.36 & 13.24 &   -   & 73.73 & 49.77 & 29.63 &   -   &    -    \\
        CNM*~\cite{zheng2022weakly}         & WS  & 60.39 & 35.43 & 15.45 &   -   & 78.13 & 55.68 & 33.33 &   -   &    -    \\  \hline
        DSCNet \cite{liu2022unsupervised}   & US  & 44.15 & 28.73 & 14.67 &  -    &   -   & 47.29 & 28.16 &   -   &  -      \\  \hline
        
        PSVL* \cite{nam2021zero}            & ZS  & 46.17 & 31.29 & 14.17 & 31.24 &   -   & 44.74 & 30.08 & 14.74 & 29.62   \\
        \bf{Ours}*                          & ZS  &\bf{52.95} &\bf{37.24} &\bf{19.33}  &\bf{36.05} &\bf{61.35} &\bf{47.61} &\bf{32.59} &\bf{15.42} &\bf{31.85}                                  \\  \hline
    \end{tabular}
    \vspace{3pt}
    \caption{Performance comparison with other methods on the Charades-STA and the ActivityNet Captions dataset.
    `Sup.' refers to supervision level: 
    WS (Weakly-supervised setting),
    US (Unsupervised setting, where query information utilized but not paired to videos),
    ZS (Zero-shot setting, where any annotation are not exploited including query information)
    * These works use pretrained models: ours and \cite{zheng2022weakly} use frozen CLIP, and \cite{nam2021zero} fine-tune RoBERTa~\cite{liu2019roberta}.}
    \label{tab:1}
\end{table*}

\subsection{Video Grounding Model}
In this section, we describe our video grounding model consisting of a video encoder and a cross-modality fusion module that learns to fuse two distinct modality features.

\paragraph{Video encoding.} 
We reuse the obtained video feature $f$ in the video grounding model with temporal positional encoding.
As our goal is to regress the temporal boundaries, it is important to embed the position information.
 
To explicitly model the position information of each video, we apply temporal positional encoding $e_{\text{pos}}$ of each segment as done in \cite{vaswani2017attention}. 
Then we apply bi-directional GRU \cite{chung2014empirical} to further encode temporal information. 
A final representation of the video $s$ is obtained by aggregating the vector that concatenates the last hidden layer of bi-directional GRU and the positional encoded video feature as follows:
\begin{equation}
    s =  \texttt{MLP}[\texttt{Bi-GRU}(\hat{f})\oplus\hat{f}],
\end{equation}
where $\oplus$ is a concatenation operation and $\hat{f}=f+e_{\text{pos}}$ is a video feature that combines positional embeddings.

\paragraph{Cross-modality fusion module.}
Given obtaining the pseudo language feature $\tilde{q}$ and the encoded whole video feature $s$, video grounding aims to find the most related parts in the video corresponding to the given language feature.
To achieve the goal, we leverage an attention mechanism proposed in \cite{vaswani2017attention} to enable the multi-modal interaction of the two modalities. 
Specifically, we obtain language-guided video feature $s_{att}$ using multi-head attention where we denote query $Q$ as the video feature $f$, key $K$ and value $V$ as the pseudo language feature $\tilde{q}$:
\begin{equation}
    \texttt{Cross-Attention}(Q,K,V) = \texttt{softmax}(\frac{QK^T}{\sqrt{d_k}}V),
\end{equation}
where $d_k$ is the dimension of $K$. 
Then, to capture more global context across the video, we additionally apply a self-attention layer after the cross-attention layer. 
We carefully note that cross-attention and self-attention have different roles in the fusion module, where key, query, and value in the self-attention layer are the video attention feature, and the key and value of the cross-attention layer is the pseudo language feature. 
Finally, with an MLP layer, we predict the start and end time stamps of the most relevant temporal region from the condensed video feature. 
This process is summarized as follows:
\begin{equation}
    (\hat{t_s}, \hat{t_e})=\texttt{MLP}(\texttt{Self-Attention}(s_{att})),
\end{equation}
where $(\hat{t_s}, \hat{t_e})$ is the predicted start and end time, respectively.

\subsection{Model Training and Inference}
Since our method performs video grounding in the zero-shot setting, the training and inference processes are different as illustrated~\figref{fig:main}.
Next, we describe training objectives to learn the video grounding model with the pseudo temporal proposal and the pseudo language query, and the inference process with the given video and real language query.

\paragraph{Training.}
Our training objective includes two loss functions, the temporal regression loss $L_{reg}$ and the temporal attention calibration loss $L_{att}$:
\begin{equation}
    L = L_{reg} + \lambda L_{att}. 
\end{equation}
To balance each objective term, the hyper-parameter $\lambda$ is used. Note that we empirically select $\lambda$ to 1, which has shown less effect on training.

Following previous works~\cite{mun2020local,yuan2019find}, we adopt temporal regression loss $L_{reg}$ as a smooth L1 loss between model prediction and target interval, which is given by
\begin{equation}
    L_{reg} = smooth_{L_1}(\hat{t_s} - \tilde{t_s}) + smooth_{L_1}(\hat{t_e} - \tilde{t_e}),
\end{equation}
where $(\tilde{t_s}, \tilde{t_e})$ and $(\hat{t_s}, \hat{t_e})$ denotes pseudo temporal ground-truth and model prediction, respectively.

We also adopt temporal attention calibration loss $ L_{att} $ to increase the accuracy of temporal attention since we directly regress the time intervals from temporally attended video features following \cite{yuan2019find}:
\begin{equation}
     L_{att} =  -\frac{\sum_{t=1}^T{\tilde{a_t} \log(a_t)}}{\sum_{t=1}^T{\tilde{a_t}}},
\end{equation}
where 
\begin{equation}
     \tilde{a_t} = \begin{cases}  1, & \mbox{if } \tilde{t_s} \leq t \leq  \tilde{t_e}\\
                                0, & otherwise.
\end{cases}
\end{equation}

\begin{table}[t]
    \centering
    \resizebox{0.95\linewidth}{!}{
    \begin{tabular}{c c  | c c c c}
        \hline
        $L_{reg}$   & $L_{att}$     & R@0.3 & R@0.5 & R@0.7 & mIoU     \\  \hline 
        \cmark      &  \xmark       & 45.16 & 30.40 & 14.88 & 30.33     \\ 
        \xmark      &  \cmark       & 12.81 & 8.71  & 3.99 & 8.71      \\
        \cmark      &  \cmark       &\bf{52.95} & \bf{37.24} & \bf{19.33}  & \bf{36.05}     \\ \hline
    \end{tabular}}
    \vspace{3pt}
    \caption{The ablation study of different losses. \cmark~means the loss term used in training.}\label{tab:3}\vspace{-11pt}
\end{table}

\paragraph{Inference.} Different from the training process, in the inference stage, an input is a video and its corresponding complete sentences from the test set. To deal with this difference, we extract text features from the text encoder of the pretrained vision-language model, \ie text encoder of CLIP~\cite{radford2021learning}.
In other words, the pseudo language feature $\tilde{q}$ is replaced with the real language feature $q$ from the real language query.
Hence, our proposal generation step in~\secref{sec:32} and pseudo language feature generation step in~\secref{sec:33} are only leveraged to train the video grounding model.

\section{Experimental Results}

\subsection{Datasets}
In order to verify the effectiveness of our method, we conduct experiments on two datasets: Charades-STA~\cite{gao2017tall} and ActivityNet Captions~\cite{krishna2017dense}. Since we formulate the video grounding task as a language-free setup, any annotations related to the videos have not been utilized while training, but in the test only.  

\paragraph{Charades-STA} Charades-STA was introduced by \cite{gao2017tall} from the Charades dataset~\cite{sigurdsson2016hollywood} with the purpose of evaluating on video grounding task by annotating in a semi-automatic way. The dataset contains 12,408/3720 segment-sentence pairs and 5338/1334 videos in training and test set, respectively.

\paragraph{ActivityNet Captions} ActivityNet Captions was originally collected by \cite{krishna2017dense} for evaluating dense video captioning, which contains 37,417/17,505/17,031 segment-sentence pairs and 10,009/4917/5044 videos in training, val\_1 and val\_2, respectively. Following previous works \cite{mun2020local,nam2021zero}, we evaluate our performance on the validation set since the annotation of the test set is unavailable.

\begin{table}[t]
    \centering
    \resizebox{0.99\linewidth}{!}{
    \begin{tabular}{ c  | c c c c}
        \hline
        Model                   & R@0.3 & R@0.5 & R@0.7 & mIoU     \\  \hline 
        Ours                    & 52.95 & 37.24 & 19.33 & 36.05    \\
        Ours + \emph{temporal GT}    & 54.00	& 39.91 & 19.46 & 36.29    \\
        \hline
    \end{tabular}}
    \vspace{3pt}
    \caption{
    Upper bound analysis using ground-truth temporal boundaries (\emph{temporal GT}). With \emph{temporal GT}, we directly generate the pseudo language features corresponding to GT time intervals.
    }\label{tab:4}
\end{table}

\begin{table}[t]
    \centering
    \resizebox{0.99\linewidth}{!}{
    \begin{tabular}{ c  | c c c c}
        \hline
        Frame Selection         & R@0.3 & R@0.5 & R@0.7 & mIoU     \\  \hline 
        Random        & 50.2  & 34.84 & 15.66 & 33.49   \\
        \texttt{ST}   & \textbf{52.95} & \textbf{37.24} & \textbf{19.33} & \textbf{36.05}    \\
        \hline
    \end{tabular}}
    \vspace{3pt}
    \caption{The ablation study of frame selection strategies. `\texttt{ST}' denotes the proposed selection transformer.}\label{tab:5}\vspace{-12pt}
\end{table}

\subsection{Evaluation Metric}
To evaluate the performance of our model, we adopt R@tIoU and mIoU (mean averaged tIoU) following previous works \cite{gao2017tall, lin2020weakly} for a fair comparison. Specifically, given predicted boundaries, we compute temporal intersection over union (tIoU) with ground-truth boundaries.
R@tIoU is the percentage of the predictions which are larger than the thresholds, \ie \{0.3, 0.5, 0.7\}. mIoU is the average IoU of all the predictions.

\subsection{Implementation Details}
For a fair comparison, we employ I3D~\cite{carreira2017quo} and C3D~\cite{tran2015learning} networks as video feature extractor for the Charades-STA and ActivityNet Captions datasets, respectively, following previous works~\cite{ mun2020local,nam2021zero}. We set the maximum length $T$ of the video features to 128 in both datasets. 
For generating pseudo language features, we use pretrained CLIP-ViT/B-32.
We set $N=9$ for frame sampling and use a low-capacity transformer~\cite{vaswani2017attention} with 2 layers and 2 attention heads for the frame selection process, making it computationally efficient.
The bidirectional GRU layers in the video encoder are 2 layers architecture with a hidden size of 256. 
For the cross-modality fusion module, we use multi-head attention with 3 layers and 4 heads. The dimension of their hidden state is 256. 
For the hyperparameters, we empirically set $k=5$, $\xi=0.0001$ and $\lambda=1$.
In all experiments, we train our models with a batch size of 256 using Adam~\cite{kingma2014adam} with a fixed learning rate of 0.0004.
We provide more details in supplementary material and the code will be publicly available soon.

\subsection{Comparisons to the State-Of-The-Art}

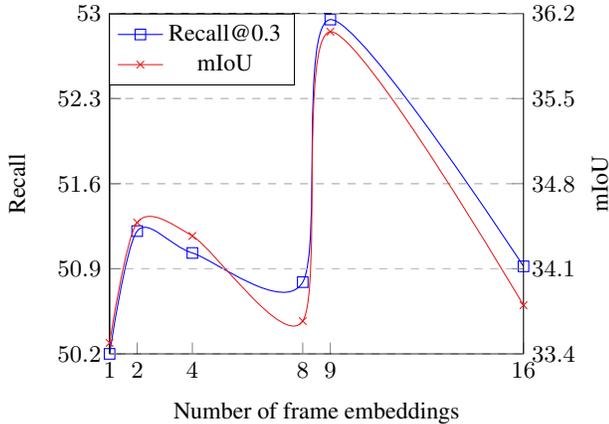
\begin{figure}
\centering
\begin{tikzpicture}{every node}=[font=\small]
\begin{axis}[
    width=0.85\linewidth, 
    axis y line*=left,
    xlabel={Number of frame embeddings},
    ylabel={Recall},
    xmin=1, xmax=16,
    ymin=50.2, ymax=53,
    xtick={1,2,4,8,9,16},
    xtick pos=left,
    ytick={50.2, 50.9,51.6,52.3, 53},
    legend pos=north west,
    ymajorgrids=true,
    xmajorgrids=false,
    grid style=dashed,
    ymajorgrids=true,
    label style = {font=\small},
    ticklabel style = {font=\small},
]

\addplot[
    color=blue,
    mark=square,
    smooth
    ]
    coordinates {
    (1,50.2)(2,51.21)(4,51.03)(8,50.79)(9,52.95)(16,50.92)
    };
    \label{Recall@0.3}

   \end{axis}
   
 \begin{axis}[
    width=0.85\linewidth, 
    yticklabel pos=right,
    axis x line=none,
    ylabel={mIoU},
    xmin=1, xmax=16,
    ymin=33.4, ymax=36.2,
    xtick={1,2,4,8,16},
    xtick pos=left,
    ytick={33.4, 34.1, 34.8, 35.5, 36.2},
    ymajorgrids=true,
    xmajorgrids=false,
    grid style=dashed,
    ylabel near ticks,
    label style = {font=\small},
    ticklabel style = {font=\small},
    legend style={
    at={(0.223, 1)},
    anchor=north, font=\small,
    fill opacity=0.8, draw opacity=1,text opacity=1},
    ]
]
\addlegendimage{/pgfplots/refstyle=Recall@0.3}\addlegendentry{Recall@0.3}
\addplot[
color=red,
mark=x,
smooth
]coordinates {
    (1,33.49)(2,34.48)(4,34.37)(8,33.67)(9,36.05)(16,33.8)
    };
    \addlegendentry{mIoU}
    
\end{axis}
\end{tikzpicture}
\caption{The ablation study of the number of frame embeddings used in frame selection process.}
\label{fig:frameembeddings_cr}\vspace{-8pt}
\end{figure}

\tabref{tab:1} shows the results of our model compared to previous works in fully-supervised, weakly-supervised, unsupervised, and zero-shot conditions. 
The weakly-supervised (WS) methods are trained with costly annotated sentence queries, whereas the unsupervised (US) method leverages the unpaired data of videos and sentence queries in the dataset.
However, zero-shot methods, including ours, leverage only videos in the dataset for training.
On both Charades-STA and ActivityNet Captions datasets, we can observe that our method outperforms PSVL~\cite{nam2021zero} in all metrics by large margins, demonstrating the robustness of the proposed method.
Furthermore, even though our method does not use a bunch of language queries of the dataset, our method outperforms the unsupervised method~\cite{liu2022unsupervised} by a large margin.
The comparisons with the weakly-supervised methods show that our method achieves comparable or even superior performance to several approaches~\cite{chen2020look,lin2020weakly,mithun2019weakly,song2020weakly,tan2021logan,wang2021weakly,wu2020reinforcement,zhang2020counterfactual}.

\subsection{Analysis}\label{sec:45}
To prove the excellence of our methods, we perform ablation studies and analysis from various perspectives on the Charades-STA.

\paragraph{Effects of different losses.}
We first investigate the effectiveness of using different loss terms, $L_{reg}$ and $L_{att}$. 
As shown in~\tabref{tab:3}, our model performs best when we used all loss terms, which demonstrates that using two losses is critical for training our network. 
We also find that the regression loss $L_{reg}$ had a more influence on overall performance, however, training with only regression loss $L_{reg}$ find to be inferior to the performance of baseline.

\paragraph{Upper bound analysis.}
In~\tabref{tab:4}, we give the upper bound analysis to our model by replacing the pseudo temporal proposals $(\tilde{t}_s, \tilde{t}_e)$ into temporal ground-truth $({t}_s, {t}_e)$. 
Replacing with temporal ground-truth leads to performance improvement which outperforms most of the existing weakly-supervised video grounding methods.
However, the gain was not significant because we obtained a pseudo language feature by selecting one of the frames within the generated temporal boundaries.
We carefully assume that the precise temporal location has a limited impact.

\paragraph{Effectiveness of the selection transformer.}

To investigate the importance of using the selection transformer in the pseudo language feature generation process, we replace it with a random selection module. 
In this analysis, we randomly sample a feature from extracted visual features in the proposal as a pseudo language feature.
As shown in~\tabref{tab:5}, we observe that using a selection transformer can boost the performance, suggesting the selection transformer's capability of selecting a dominant feature.

\paragraph{Effect of the number of frame embeddings.}

\begin{figure}[t]
    \centering
    {\includegraphics[width=0.85\linewidth]{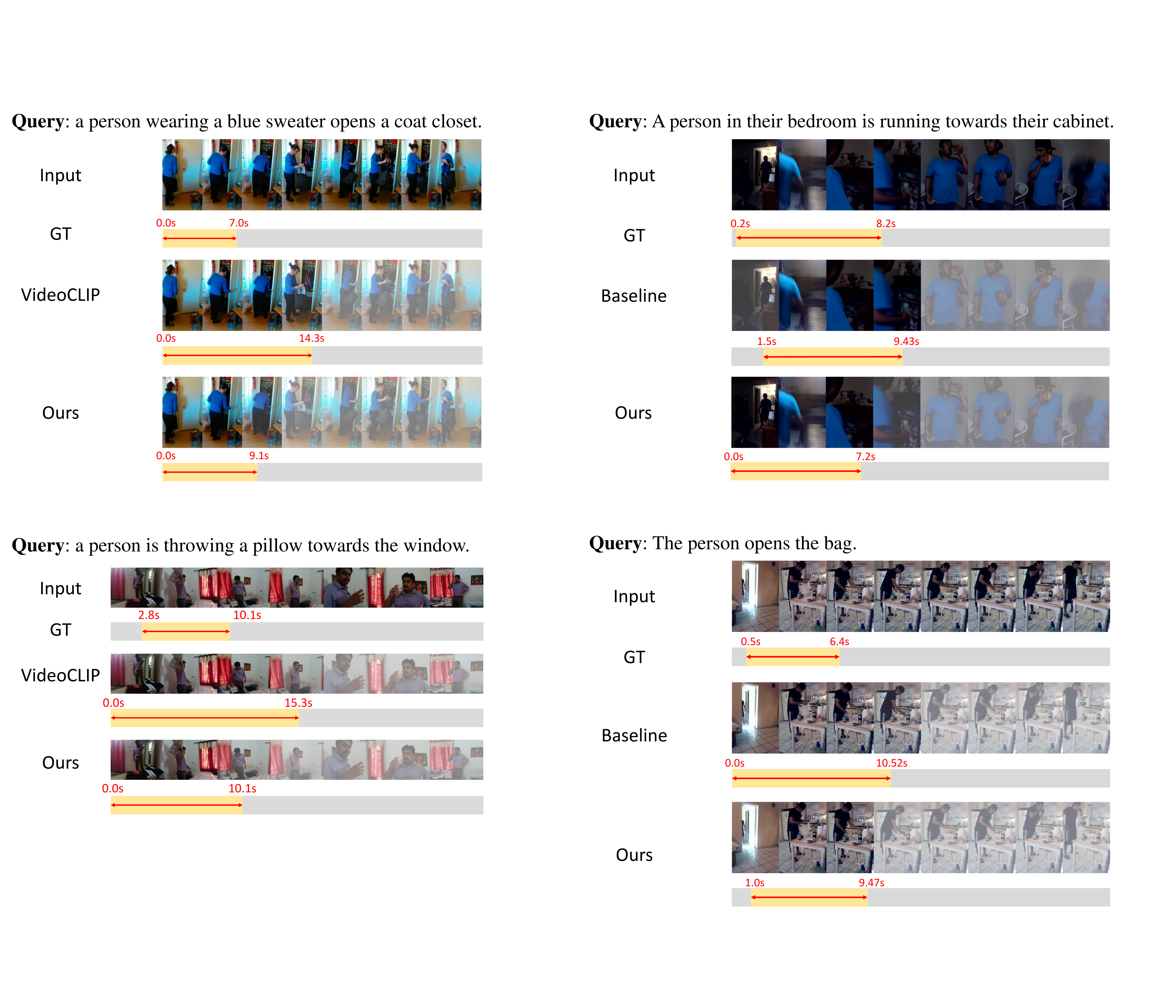}\label{fig:qual2-a}}\hfill \\ 
    \vspace{2.2pt}
    {\includegraphics[width=0.85\linewidth]{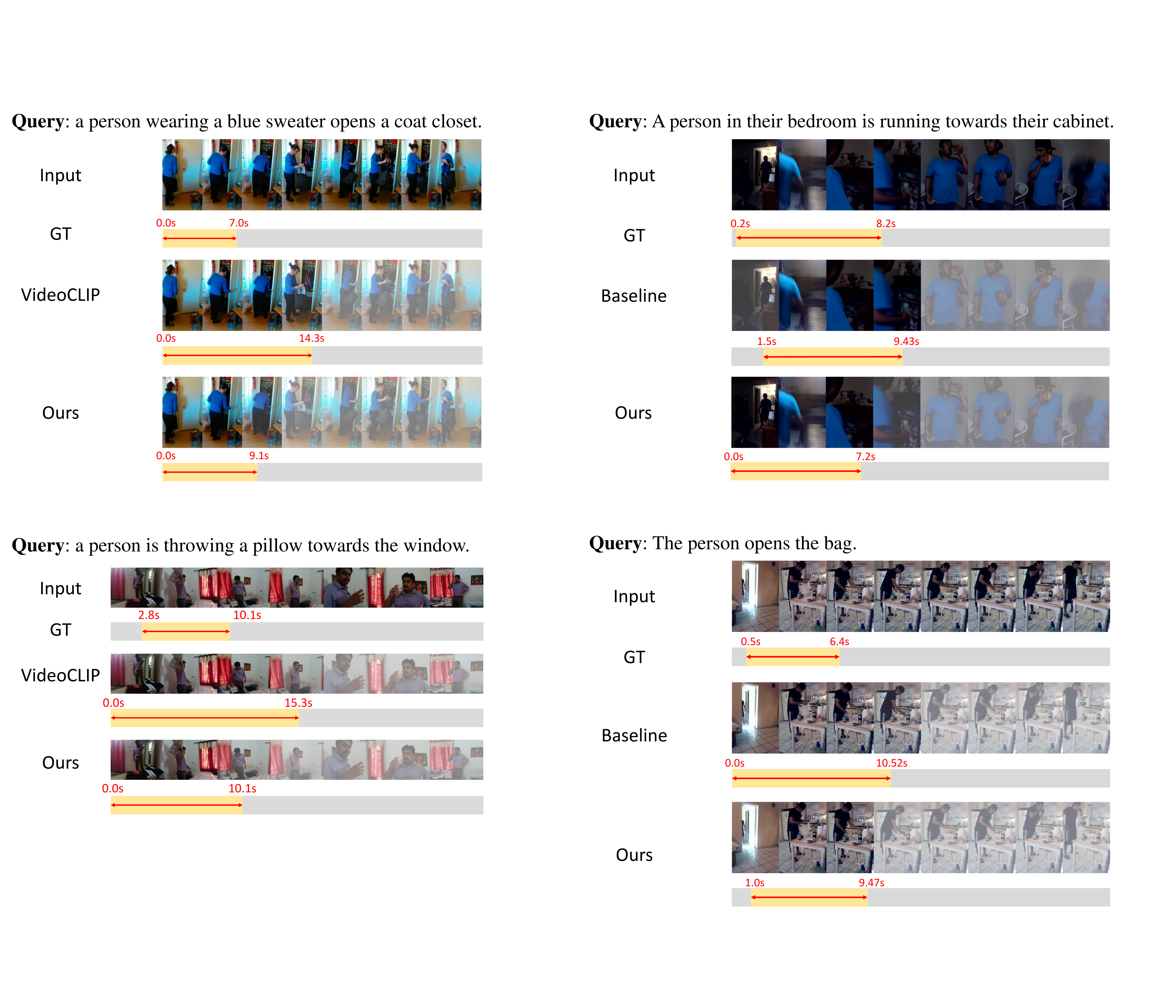}\label{fig:qual2-b}}\hfill  \\
    \caption{
    Qualitative comparisons corresponding to the language feature encoders on the Charades-STA dataset.
    }\label{fig:qual2}\vspace{-10pt}
\end{figure}

As shown in~\figref{fig:frameembeddings_cr}, we evaluate the effectiveness of the number of frame embeddings used in the selection transformer. We can see that the more we sample the frame embeddings, the higher the tIoU scores until the 9 frames. So, we set the $N=9$ in all experiments. More results for the recall at different tIoU are shown in the supplementary material.

\paragraph{Effectiveness of the image-based vision-language model.}

In this section, we investigate the effectiveness of the image-based vision-language model for the video-language task. 
For this experiment, we employ pretrained video-language model~\cite{xu2021videoclip}, which has established fine-grained associations between video and text with contrastive loss, to generate a pseudo language feature.
The pseudo language feature is directly obtained from the video-language model extracted from the proposal. 
\figref{fig:qual2} shows some qualitative results for the comparison between our method with CLIP and with its counterpart (i.e., VideoCLIP).
As shown in~\figref{fig:qual2}, our model with CLIP can localize the better moment than with VideoCLIP, regardless of whether the given query is more static or dynamic.
We observe that using an image-language model can capture semantic information from a single frame comparable to or better than the video-language model.

\subsection{Qualitative Results}

\begin{figure}[t]
    \centering
    {\includegraphics[width=0.87\linewidth]{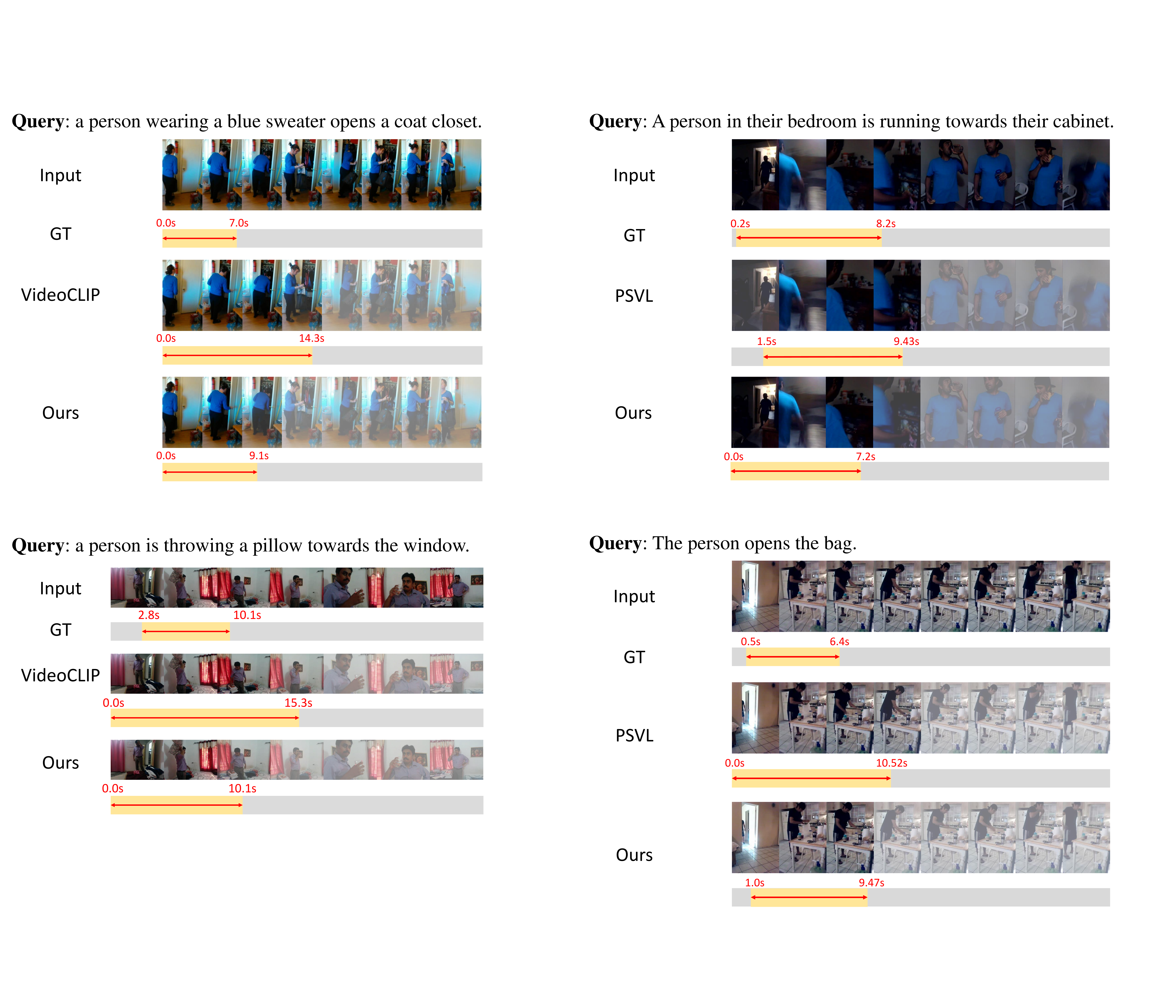}\label{fig:qual-a}}\hfill \\ 
    \vspace{2.2pt}
    {\includegraphics[width=0.87\linewidth]{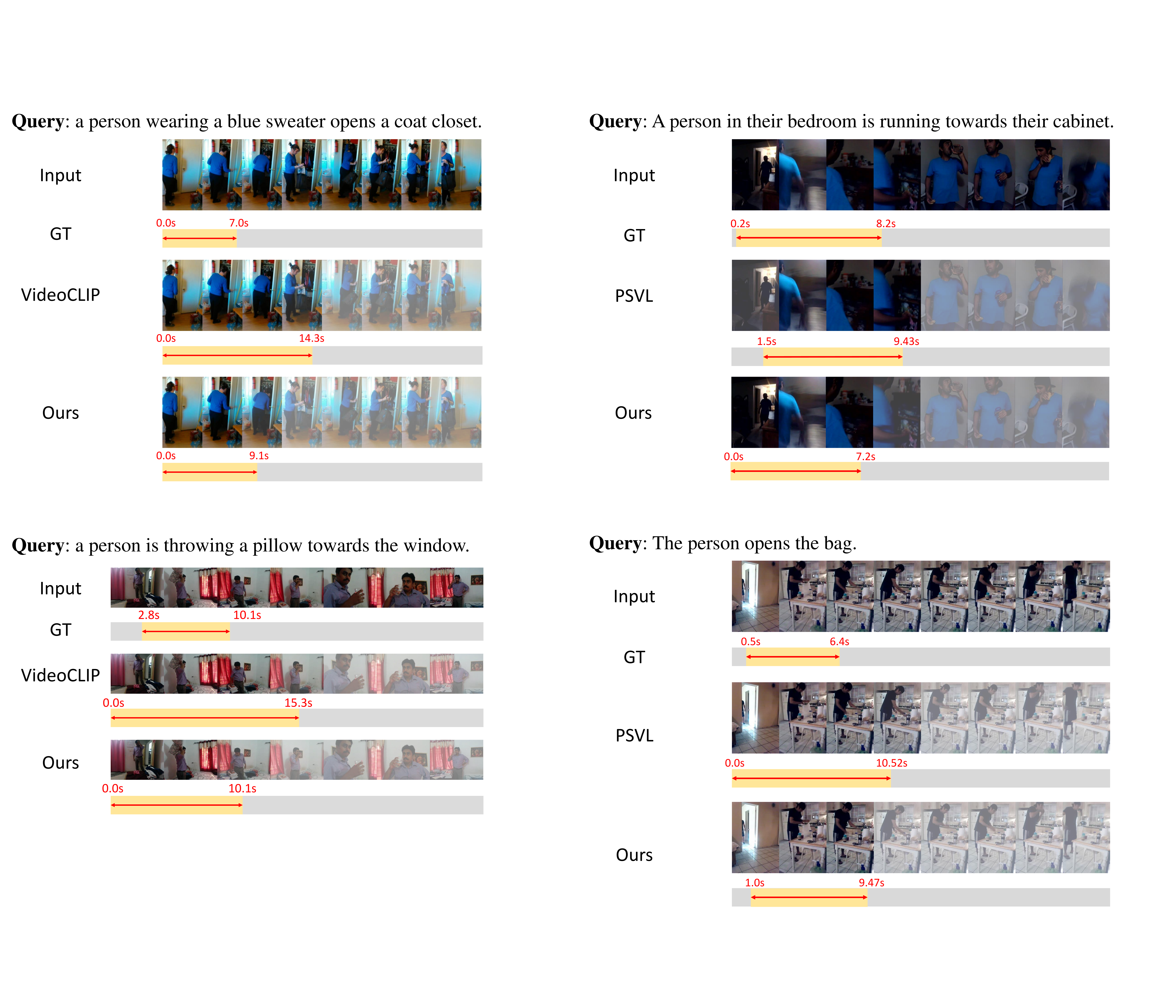}\label{fig:qual-b}}\hfill  \\ 
    \caption{
    Qualitative comparisons between ours and PSVL on the Charades-STA dataset.
    }\label{fig:qual}\vspace{-10pt}
\end{figure}

\begin{figure}[t]
    \centering
    {\includegraphics[width=1.0\linewidth]{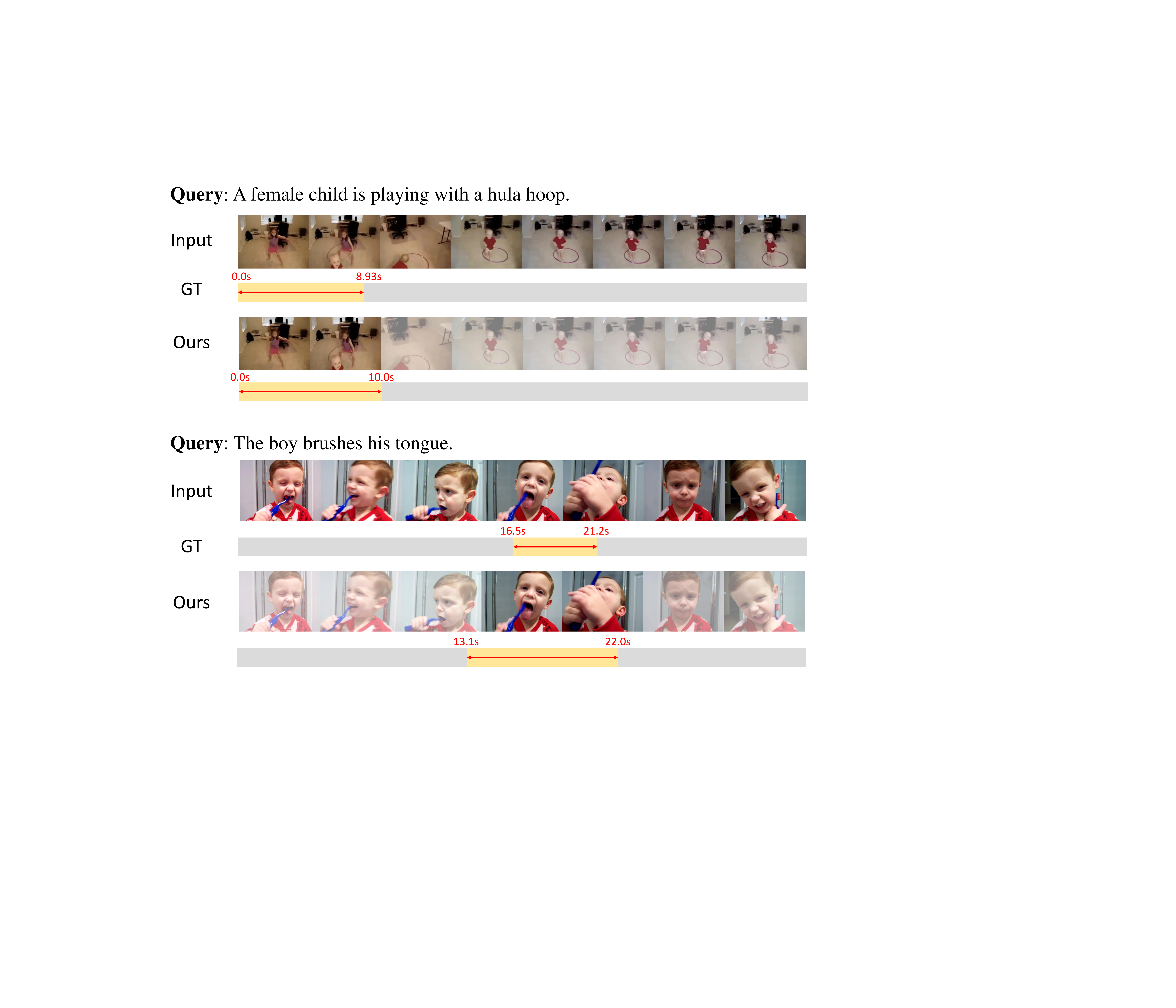}\label{fig:qual3-a}}\hfill \\ 
    \vspace{2.2pt}
    {\includegraphics[width=1.0\linewidth]{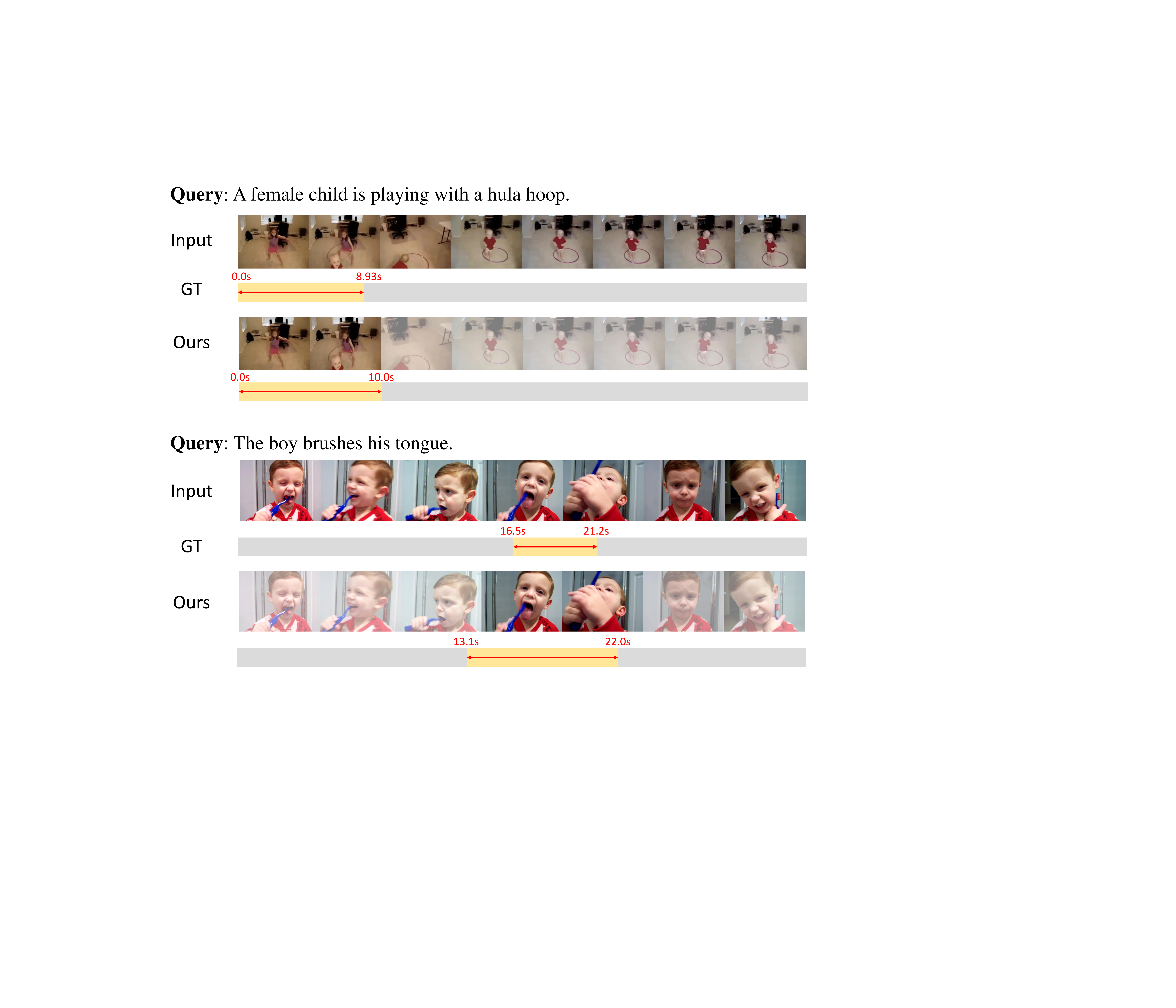}\label{fig:qual3-b}}\hfill  \\ 
    \caption{Qualitative comparisons between ground-truth intervals and ours on the ActivityNet Captions dataset.
    }\label{fig:qual3}\vspace{-10pt}
\end{figure}

\figref{fig:qual} shows some qualitative results comparing our results with the previous method~\cite{nam2021zero} on the Charades-STA dataset.
This example presents the temporal ground-truth boundaries and model predictions of PSVL~\cite{nam2021zero} and ours, given a pair of a video and a query.
The results show that the proposed method covers more of the video content related to the query, which effectively shows that our model is qualitatively better than PSVL. 
Also,~\figref{fig:qual3} illustrates the qualitative results on the ActivityNet Captions dataset. The more qualitative results are in the supplementary material.

\section{Conclusion and Future Work}

In this work, we present a novel method to train a video grounding model in a zero-shot manner without using any annotation related to paired video-sentence data.
We achieve the goal by generating pseudo ground-truth of temporal locations and corresponding text features with the language-free paradigm.
Primarily, we obtain a pseudo language feature from a generated proposal leveraging the well-aligned visual-language semantic space of CLIP.
In contrast to the previous method of trying to make pseudo text queries into contrived language formats, we preserve the structural characteristics and compositional generalization inherent in natural language.
Moreover, we develop a video grounding model based on cross- and self-attention transformers to effectively model the relationship between two modalities and the context of attended features.
The experimental results demonstrate the efficacy of language-free training, achieving remarkable performances on two datasets and reducing the cost of data collection.

However, the temporal modeling is not designed in this work due to the aforementioned reasons in~\secref{sec:33}. As shown in the experimental results, current datasets show a limitation for temporal reasoning in the settled query. As the next step, we will investigate the new benchmark for video grounding, which should include more hard examples of causal and temporal understanding as well as more long-term videos for practical usage. 

\paragraph{Acknowledgements.}
This work was supported by the Yonsei Signature Research Cluster Program of 2022 (2022-22-0002) and the KIST Institutional Program (Project No.2E31051-21-203).

{\small
\bibliographystyle{ieee_fullname}
\bibliography{egbib}

\begin{thebibliography}{10}\itemsep=-1pt

\bibitem{anne2017localizing}
Lisa Anne~Hendricks, Oliver Wang, Eli Shechtman, Josef Sivic, Trevor Darrell,
  and Bryan Russell.
\newblock Localizing moments in video with natural language.
\newblock {\em ICCV}, 2017.

\bibitem{buch2022revisiting}
Shyamal Buch, Crist{\'o}bal Eyzaguirre, Adrien Gaidon, Jiajun Wu, Li Fei-Fei,
  and Juan~Carlos Niebles.
\newblock Revisiting the ``video" in video-language understanding.
\newblock {\em CVPR}, 2022.

\bibitem{carreira2017quo}
Joao Carreira and Andrew Zisserman.
\newblock Quo vadis, action recognition? a new model and the kinetics dataset.
\newblock {\em CVPR}, 2017.

\bibitem{chen2020look}
Zhenfang Chen, Lin Ma, Wenhan Luo, Peng Tang, and Kwan-Yee~K Wong.
\newblock Look closer to ground better: Weakly-supervised temporal grounding of
  sentence in video.
\newblock {\em arXiv preprint arXiv:2001.09308}, 2020.

\bibitem{chomsky2009syntactic}
Noam Chomsky.
\newblock Syntactic structures.
\newblock {\em De Gruyter Mouton}, 2009.

\bibitem{chung2014empirical}
Junyoung Chung, Caglar Gulcehre, KyungHyun Cho, and Yoshua Bengio.
\newblock Empirical evaluation of gated recurrent neural networks on sequence
  modeling.
\newblock {\em arXiv preprint arXiv:1412.3555}, 2014.

\bibitem{collins2000system}
Robert~T Collins, Alan~J Lipton, Takeo Kanade, Hironobu Fujiyoshi, David
  Duggins, Yanghai Tsin, David Tolliver, Nobuyoshi Enomoto, Osamu Hasegawa,
  Peter Burt, et~al.
\newblock A system for video surveillance and monitoring.
\newblock {\em VSAM final report}, 2000.

\bibitem{dietterich1997solving}
Thomas~G Dietterich, Richard~H Lathrop, and Tom{\'a}s Lozano-P{\'e}rez.
\newblock Solving the multiple instance problem with axis-parallel rectangles.
\newblock {\em Artificial intelligence}, 1997.

\bibitem{ding2021support}
Xinpeng Ding, Nannan Wang, Shiwei Zhang, De Cheng, Xiaomeng Li, Ziyuan Huang,
  Mingqian Tang, and Xinbo Gao.
\newblock Support-set based cross-supervision for video grounding.
\newblock {\em ICCV}, 2021.

\bibitem{duan2018weakly}
Xuguang Duan, Wenbing Huang, Chuang Gan, Jingdong Wang, Wenwu Zhu, and Junzhou
  Huang.
\newblock Weakly supervised dense event captioning in videos.
\newblock {\em NeurIPS}, 2018.

\bibitem{dwibedi2020counting}
Debidatta Dwibedi, Yusuf Aytar, Jonathan Tompson, Pierre Sermanet, and Andrew
  Zisserman.
\newblock Counting out time: Class agnostic video repetition counting in the
  wild.
\newblock {\em CVPR}, 2020.

\bibitem{feng2019unsupervised}
Yang Feng, Lin Ma, Wei Liu, and Jiebo Luo.
\newblock Unsupervised image captioning.
\newblock {\em CVPR}, 2019.

\bibitem{fodor1988connectionism}
Jerry~A Fodor and Zenon~W Pylyshyn.
\newblock Connectionism and cognitive architecture: A critical analysis.
\newblock {\em Cognition}, 1988.

\bibitem{gao2017tall}
Jiyang Gao, Chen Sun, Zhenheng Yang, and Ram Nevatia.
\newblock Tall: Temporal activity localization via language query.
\newblock {\em ICCV}, 2017.

\bibitem{gao2019wslln}
Mingfei Gao, Larry~S Davis, Richard Socher, and Caiming Xiong.
\newblock Wslln: Weakly supervised natural language localization networks.
\newblock {\em EMNLP}, 2019.

\bibitem{huang2021cross}
Jiabo Huang, Yang Liu, Shaogang Gong, and Hailin Jin.
\newblock Cross-sentence temporal and semantic relations in video activity
  localisation.
\newblock {\em ICCV}, 2021.

\bibitem{jiang2022pseudo}
Haojun Jiang, Yuanze Lin, Dongchen Han, Shiji Song, and Gao Huang.
\newblock Pseudo-q: Generating pseudo language queries for visual grounding.
\newblock {\em CVPR}, 2022.

\bibitem{kingma2014adam}
Diederik~P Kingma and Jimmy Ba.
\newblock Adam: A method for stochastic optimization.
\newblock {\em ICLR}, 2015.

\bibitem{krishna2017dense}
Ranjay Krishna, Kenji Hata, Frederic Ren, Li Fei-Fei, and Juan Carlos~Niebles.
\newblock Dense-captioning events in videos.
\newblock {\em ICCV}, 2017.

\bibitem{laina2019towards}
Iro Laina, Christian Rupprecht, and Nassir Navab.
\newblock Towards unsupervised image captioning with shared multimodal
  embeddings.
\newblock {\em ICCV}, 2019.

\bibitem{lei2021less}
Jie Lei, Linjie Li, Luowei Zhou, Zhe Gan, Tamara~L Berg, Mohit Bansal, and
  Jingjing Liu.
\newblock Less is more: Clipbert for video-and-language learning via sparse
  sampling.
\newblock {\em CVPR}, 2021.

\bibitem{li2022compositional}
Juncheng Li, Junlin Xie, Long Qian, Linchao Zhu, Siliang Tang, Fei Wu, Yi Yang,
  Yueting Zhuang, and Xin~Eric Wang.
\newblock Compositional temporal grounding with structured variational
  cross-graph correspondence learning.
\newblock {\em CVPR}, 2022.

\bibitem{lin2020weakly}
Zhijie Lin, Zhou Zhao, Zhu Zhang, Qi Wang, and Huasheng Liu.
\newblock Weakly-supervised video moment retrieval via semantic completion
  network.
\newblock {\em AAAI}, 2020.

\bibitem{liu2021context}
Daizong Liu, Xiaoye Qu, Jianfeng Dong, Pan Zhou, Yu Cheng, Wei Wei, Zichuan Xu,
  and Yulai Xie.
\newblock Context-aware biaffine localizing network for temporal sentence
  grounding.
\newblock {\em CVPR}, 2021.

\bibitem{liu2022unsupervised}
Daizong Liu, Xiaoye Qu, Yinzhen Wang, Xing Di, Kai Zou, Yu Cheng, Zichuan Xu,
  and Pan Zhou.
\newblock Unsupervised temporal video grounding with deep semantic clustering.
\newblock {\em AAAI}, 2022.

\bibitem{liu2019roberta}
Yinhan Liu, Myle Ott, Naman Goyal, Jingfei Du, Mandar Joshi, Danqi Chen, Omer
  Levy, Mike Lewis, Luke Zettlemoyer, and Veselin Stoyanov.
\newblock Roberta: A robustly optimized bert pretraining approach.
\newblock {\em arXiv preprint arXiv:1907.11692}, 2019.

\bibitem{ma2020vlanet}
Minuk Ma, Sunjae Yoon, Junyeong Kim, Youngjoon Lee, Sunghun Kang, and Chang~D
  Yoo.
\newblock Vlanet: Video-language alignment network for weakly-supervised video
  moment retrieval.
\newblock {\em ECCV}, 2020.

\bibitem{mithun2019weakly}
Niluthpol~Chowdhury Mithun, Sujoy Paul, and Amit~K Roy-Chowdhury.
\newblock Weakly supervised video moment retrieval from text queries.
\newblock {\em CVPR}, 2019.

\bibitem{mun2020local}
Jonghwan Mun, Minsu Cho, and Bohyung Han.
\newblock Local-global video-text interactions for temporal grounding.
\newblock {\em CVPR}, 2020.

\bibitem{nam2021zero}
Jinwoo Nam, Daechul Ahn, Dongyeop Kang, Seong~Jong Ha, and Jonghyun Choi.
\newblock Zero-shot natural language video localization.
\newblock {\em ICCV}, 2021.

\bibitem{park2020sum}
Jungin Park, Jiyoung Lee, Ig-Jae Kim, and Kwanghoon Sohn.
\newblock Sumgraph: Video summarization via recursive graph modeling.
\newblock {\em ECCV}, 2020.

\bibitem{park2021b2a}
Jungin Park, Jiyoung Lee, and Kwanghoon Sohn.
\newblock Bridge to answer: Structure-aware graph interaction networks for
  video question answering.
\newblock {\em CVPR}, 2021.

\bibitem{radford2021learning}
Alec Radford, Jong~Wook Kim, Chris Hallacy, Aditya Ramesh, Gabriel Goh,
  Sandhini Agarwal, Girish Sastry, Amanda Askell, Pamela Mishkin, Jack Clark,
  et~al.
\newblock Learning transferable visual models from natural language
  supervision.
\newblock {\em ICML}, 2021.

\bibitem{rodriguez2020proposal}
Cristian Rodriguez, Edison Marrese-Taylor, Fatemeh~Sadat Saleh, Hongdong Li,
  and Stephen Gould.
\newblock Proposal-free temporal moment localization of a natural-language
  query in video using guided attention.
\newblock {\em WACV}, 2020.

\bibitem{schuhmann2022laion}
Christoph Schuhmann, Romain Beaumont, Richard Vencu, Cade Gordon, Ross
  Wightman, Mehdi Cherti, Theo Coombes, Aarush Katta, Clayton Mullis, Mitchell
  Wortsman, et~al.
\newblock Laion-5b: An open large-scale dataset for training next generation
  image-text models.
\newblock {\em arXiv preprint arXiv:2210.08402}, 2022.

\bibitem{sigurdsson2016hollywood}
Gunnar~A Sigurdsson, G{\"u}l Varol, Xiaolong Wang, Ali Farhadi, Ivan Laptev,
  and Abhinav Gupta.
\newblock Hollywood in homes: Crowdsourcing data collection for activity
  understanding.
\newblock {\em ECCV}, 2016.

\bibitem{sivic2003video}
Josef Sivic and Andrew Zisserman.
\newblock Video google: A text retrieval approach to object matching in videos.
\newblock {\em ICCV}, 2003.

\bibitem{snoek2009mediamill}
Cees Snoek, Kvd Sande, OD Rooij, Bouke Huurnink, J Uijlings, M~van Liempt, M
  Bugalhoy, I Trancosoy, F Yan, M Tahir, et~al.
\newblock The mediamill trecvid 2009 semantic video search engine.
\newblock {\em TRECVID workshop}, 2009.

\bibitem{soldan2021vlg}
Mattia Soldan, Mengmeng Xu, Sisi Qu, Jesper Tegner, and Bernard Ghanem.
\newblock Vlg-net: Video-language graph matching network for video grounding.
\newblock {\em ICCV}, 2021.

\bibitem{song2020weakly}
Yijun Song, Jingwen Wang, Lin Ma, Zhou Yu, and Jun Yu.
\newblock Weakly-supervised multi-level attentional reconstruction network for
  grounding textual queries in videos.
\newblock {\em arXiv preprint arXiv:2003.07048}, 2020.

\bibitem{tan2021logan}
Reuben Tan, Huijuan Xu, Kate Saenko, and Bryan~A Plummer.
\newblock Logan: Latent graph co-attention network for weakly-supervised video
  moment retrieval.
\newblock {\em WACV}, 2021.

\bibitem{tran2015learning}
Du Tran, Lubomir Bourdev, Rob Fergus, Lorenzo Torresani, and Manohar Paluri.
\newblock Learning spatiotemporal features with 3d convolutional networks.
\newblock {\em ICCV}, 2015.

\bibitem{vaswani2017attention}
Ashish Vaswani, Noam Shazeer, Niki Parmar, Jakob Uszkoreit, Llion Jones,
  Aidan~N Gomez, {\L}ukasz Kaiser, and Illia Polosukhin.
\newblock Attention is all you need.
\newblock {\em NeurIPS}, 2017.

\bibitem{wang2021structured}
Hao Wang, Zheng-Jun Zha, Liang Li, Dong Liu, and Jiebo Luo.
\newblock Structured multi-level interaction network for video moment
  localization via language query.
\newblock {\em CVPR}, 2021.

\bibitem{wang2021weakly}
Yuechen Wang, Jiajun Deng, Wengang Zhou, and Houqiang Li.
\newblock Weakly supervised temporal adjacent network for language grounding.
\newblock {\em IEEE TMM}, 2021.

\bibitem{wang2021visual}
Zheng Wang, Jingjing Chen, and Yu-Gang Jiang.
\newblock Visual co-occurrence alignment learning for weakly-supervised video
  moment retrieval.
\newblock {\em ACM MM}, 2021.

\bibitem{wang2022clip}
Zihao Wang, Wei Liu, Qian He, Xinglong Wu, and Zili Yi.
\newblock Clip-gen: Language-free training of a text-to-image generator with
  clip.
\newblock {\em arXiv preprint arXiv:2203.00386}, 2022.

\bibitem{wu2020reinforcement}
Jie Wu, Guanbin Li, Xiaoguang Han, and Liang Lin.
\newblock Reinforcement learning for weakly supervised temporal grounding of
  natural language in untrimmed videos.
\newblock {\em ACM MM}, 2020.

\bibitem{xu2021videoclip}
Hu Xu, Gargi Ghosh, Po-Yao Huang, Dmytro Okhonko, Armen Aghajanyan, Florian
  Metze, Luke Zettlemoyer, and Christoph Feichtenhofer.
\newblock Videoclip: Contrastive pre-training for zero-shot video-text
  understanding.
\newblock {\em EMNLP}, 2021.

\bibitem{yang2021local}
Wenfei Yang, Tianzhu Zhang, Yongdong Zhang, and Feng Wu.
\newblock Local correspondence network for weakly supervised temporal sentence
  grounding.
\newblock {\em IEEE TIP}, 2021.

\bibitem{yuan2019find}
Yitian Yuan, Tao Mei, and Wenwu Zhu.
\newblock To find where you talk: Temporal sentence localization in video with
  attention based location regression.
\newblock {\em AAAI}, 2019.

\bibitem{zeng2020dense}
Runhao Zeng, Haoming Xu, Wenbing Huang, Peihao Chen, Mingkui Tan, and Chuang
  Gan.
\newblock Dense regression network for video grounding.
\newblock {\em CVPR}, 2020.

\bibitem{zhang2019man}
Da Zhang, Xiyang Dai, Xin Wang, Yuan-Fang Wang, and Larry~S Davis.
\newblock Man: Moment alignment network for natural language moment retrieval
  via iterative graph adjustment.
\newblock {\em CVPR}, 2019.

\bibitem{zhang2020learning}
Songyang Zhang, Houwen Peng, Jianlong Fu, and Jiebo Luo.
\newblock Learning 2d temporal adjacent networks for moment localization with
  natural language.
\newblock {\em AAAI}, 2020.

\bibitem{zhang2020regularized}
Zhu Zhang, Zhijie Lin, Zhou Zhao, Jieming Zhu, and Xiuqiang He.
\newblock Regularized two-branch proposal networks for weakly-supervised moment
  retrieval in videos.
\newblock {\em ACM MM}, 2020.

\bibitem{zhang2020counterfactual}
Zhu Zhang, Zhou Zhao, Zhijie Lin, Xiuqiang He, et~al.
\newblock Counterfactual contrastive learning for weakly-supervised
  vision-language grounding.
\newblock {\em NeurIPS}, 2020.

\bibitem{zhao2021cascaded}
Yang Zhao, Zhou Zhao, Zhu Zhang, and Zhijie Lin.
\newblock Cascaded prediction network via segment tree for temporal video
  grounding.
\newblock {\em CVPR}, 2021.

\bibitem{zheng2022weakly}
Minghang Zheng, Yanjie Huang, Qingchao Chen, and Yang Liu.
\newblock Weakly supervised video moment localization with contrastive negative
  sample mining.
\newblock {\em AAAI}, 2022.

\bibitem{zhou2021embracing}
Hao Zhou, Chongyang Zhang, Yan Luo, Yanjun Chen, and Chuanping Hu.
\newblock Embracing uncertainty: Decoupling and de-bias for robust temporal
  grounding.
\newblock {\em CVPR}, 2021.

\bibitem{zhou2021lafite}
Yufan Zhou, Ruiyi Zhang, Changyou Chen, Chunyuan Li, Chris Tensmeyer, Tong Yu,
  Jiuxiang Gu, Jinhui Xu, and Tong Sun.
\newblock Lafite: Towards language-free training for text-to-image generation.
\newblock {\em CVPR}, 2022.

\bibitem{zhu2022prompt}
Peipei Zhu, Xiao Wang, Lin Zhu, Zhenglong Sun, Weishi Zheng, Yaowei Wang, and
  Changwen Chen.
\newblock Prompt-based learning for unpaired image captioning.
\newblock {\em arXiv preprint arXiv:2205.13125}, 2022.

\end{thebibliography}
}

\end{document}